\documentclass[10pt,twocolumn,letterpaper]{article}

\usepackage{cvpr}
\usepackage{times}
\usepackage{epsfig}
\usepackage{graphicx}
\usepackage{amsmath}
\usepackage{amssymb}
\usepackage{enumitem}
\usepackage{subcaption}
\usepackage{booktabs}
\usepackage{multirow}
\usepackage{pifont}
\newcommand{\cmark}{\ding{51}}
\newcommand{\xmark}{\ding{55}}

\makeatletter
\usepackage{xspace}
\def\@onedot{\ifx\@let@token.\else.\null\fi\xspace}
\DeclareRobustCommand\onedot{\futurelet\@let@token\@onedot}

\newcommand{\figref}[1]{Fig\onedot~\ref{#1}}
\newcommand{\equref}[1]{Eq\onedot~\eqref{#1}}
\newcommand{\secref}[1]{Sec\onedot~\ref{#1}}
\newcommand{\tabref}[1]{Tab\onedot~\ref{#1}}

\def\eg{\emph{e.g}\onedot} 
 
\def\cf{\emph{cf}\onedot} 
\def\etc{\emph{etc}\onedot}

\newcommand{\eat}[1]{}

\usepackage[pagebackref=true,breaklinks=true,letterpaper=true,colorlinks,bookmarks=false]{hyperref}

\cvprfinalcopy 

\def\cvprPaperID{1183} 

\ifcvprfinal\fi
\begin{document}

\title{Auto-DeepLab: \\Hierarchical Neural Architecture Search for Semantic Image Segmentation}


\author{Chenxi Liu$^1$\thanks{Work done while an intern at Google.}, Liang-Chieh Chen$^2$, Florian Schroff$^2$, Hartwig Adam$^2$, Wei Hua$^2$,\\ Alan Yuille$^1$, Li Fei-Fei$^3$\\
$^1$Johns Hopkins University \quad $^2$Google \quad $^3$Stanford University\\
}

\maketitle

\begin{abstract}
Recently, Neural Architecture Search (NAS) has successfully identified neural network architectures that exceed human designed ones on large-scale image classification. In this paper, we study NAS for semantic image segmentation. Existing works often focus on searching the repeatable cell structure, while hand-designing the outer network structure that controls the spatial resolution changes. This choice simplifies the search space, but becomes increasingly problematic for dense image prediction which exhibits a lot more network level architectural variations. Therefore, we propose to search the network level structure in addition to the cell level structure, which forms a hierarchical architecture search space. We present a network level search space that includes many popular designs, and develop a formulation that allows efficient gradient-based architecture search (3 P100 GPU days on Cityscapes images). We demonstrate the effectiveness of the proposed method on the challenging Cityscapes, PASCAL VOC 2012, and ADE20K datasets. Auto-DeepLab, our architecture searched specifically for semantic image segmentation, attains state-of-the-art performance without any ImageNet pretraining.\footnote{Code for Auto-DeepLab released at \url{https://github.com/tensorflow/models/tree/master/research/deeplab}.}
\end{abstract}

\vspace{-0.8\baselineskip}
\section{Introduction}
\vspace{-0.2\baselineskip}
\label{sec:intro}

Deep neural networks have been proved successful across a large variety of artificial intelligence tasks, including image recognition \cite{krizhevsky2012imagenet, he2016deep}, speech recognition \cite{hinton2012deep}, machine translation \cite{sutskever2014sequence, wu2016google} \etc.
While better optimizers \cite{KingmaB14} and better normalization techniques \cite{ioffe2015batch, WuH18} certainly played an important role, a lot of the progress comes from the design of neural network architectures. 
In computer vision, this holds true for both image classification \cite{krizhevsky2012imagenet, simonyan2014very, szegedy2015going, szegedy2016rethinking, szegedy2017inception, he2016deep, xie2017aggreated, huang2017densely, hu2018squeeze} and dense image prediction \cite{farabet2013learning, long2014fully, chen2014semantic, ronneberger2015u, noh2015learning, newell2016stacked}.

\begin{table}[t]
    \centering
    \scalebox{0.8}{
    \begin{tabular}{l | c c c c | c}
        \toprule[0.2em]
        & \multicolumn{4}{c|}{{ Auto Search}} \\
        { Model} & { Cell} & { Network} & { Dataset} & { Days} & { Task} \\
        \toprule[0.2em]
        ResNet \cite{he2016deep} & \xmark & \xmark & - & - & Cls \\
        DenseNet \cite{huang2017densely} & \xmark & \xmark & - & - & Cls \\
        DeepLabv3+ \cite{deeplabv3plus2018} & \xmark & \xmark & - & - & Seg \\
        \midrule
        NASNet \cite{zoph2017learning} & \cmark & \xmark & CIFAR-10 & $2000$ & Cls \\
        AmoebaNet \cite{real2018regularized} & \cmark & \xmark & CIFAR-10 & $2000$ & Cls \\
        PNASNet \cite{liu2018progressive} & \cmark & \xmark  & CIFAR-10 & $150$ & Cls \\
        DARTS \cite{liu2018darts} & \cmark & \xmark & CIFAR-10 & $4$ & Cls \\
        DPC \cite{chen2018searching} & \cmark & \xmark & Cityscapes & $2600$ & Seg \\
        \midrule 
        \midrule
        Auto-DeepLab & \cmark & \cmark & Cityscapes & $3$ & Seg \\
        \bottomrule[0.1em]
    \end{tabular}
    }
    \caption{Comparing our work against other CNN architectures with two-level hierarchy. The main differences include: (1) we directly search CNN architecture for semantic segmentation, (2) we search the network level architecture as well as the cell level one, and (3) our efficient search only requires $3$ P100 GPU days.}
    \label{tab:overview}
\end{table}

More recently, in the spirit of AutoML and democratizing AI, there has been significant interest in designing neural network architectures \emph{automatically}, instead of relying heavily on expert experience and knowledge.
Importantly, in the past year, Neural Architecture Search (NAS) has successfully identified architectures that exceed human-designed architectures on large-scale image classification problems \cite{zoph2017learning, liu2018progressive, real2018regularized}.

Image classification is a good starting point for NAS, because it is the most fundamental and well-studied high-level recognition task.
In addition, there exists benchmark datasets (\eg, CIFAR-10) with relatively small images, resulting in less computation and faster training. 
However, image classification should not be the end point for NAS, and the current success shows promise to extend into more demanding domains.
In this paper, we study Neural Architecture Search for semantic image segmentation, an important computer vision task that assigns a label like ``person'' or ``bicycle'' to each pixel in the input image.

Naively porting ideas from image classification would not suffice for semantic segmentation. In image classification, NAS typically applies  transfer learning from low resolution images to high resolution images \cite{zoph2017learning}, whereas optimal architectures for semantic  segmentation must inherently operate on high resolution imagery.
This suggests the need for: (1) a more relaxed and general search space to capture the architectural variations brought by the higher resolution, and (2) a more efficient architecture search technique as higher resolution requires heavier computation.


We notice that modern CNN designs \cite{he2016deep, xie2017aggreated, huang2017densely} usually follow a two-level hierarchy, where the outer network level controls the spatial resolution changes, and the inner cell level governs the specific layer-wise computations. 
The vast majority of current works on NAS \cite{zoph2017learning, liu2018progressive, real2018regularized, pham2018efficient, liu2018darts} follow this two-level hierarchical design, but only automatically search the inner cell level while hand-designing the outer network level.
This limited search space becomes problematic for dense image prediction, which is sensitive to the spatial resolution changes.
Therefore in our work, we propose a trellis-like network level search space that augments the commonly-used cell level search space first proposed in \cite{zoph2017learning} to form a \emph{hierarchical} architecture search space.
Our goal is to jointly learn a good combination of repeatable cell structure and network structure specifically for semantic image segmentation.

In terms of the architecture search method, reinforcement learning \cite{zoph2017neural, zoph2017learning} and evolutionary algorithms \cite{real2017large, real2018regularized} tend to be computationally intensive even on the low resolution CIFAR-10 dataset, therefore probably not suitable for semantic image segmentation.
We draw inspiration from the differentiable formulation of NAS \cite{shin2018differentiable, liu2018darts}, and develop a continuous relaxation of the discrete architectures that exactly matches the hierarchical architecture search space.
The hierarchical architecture search is conducted via stochastic gradient descent. 
When the search terminates, the best cell architecture is decoded greedily, and the best network architecture is decoded efficiently using the Viterbi algorithm. 
We directly search architecture on $321 \times 321$ image crops from Cityscapes \cite{Cordts2016Cityscapes}.
The search is very efficient and only takes about $3$ days on one P100 GPU.

We report experimental results on multiple semantic segmentation benchmarks, including Cityscapes \cite{Cordts2016Cityscapes}, PASCAL VOC 2012 \cite{everingham2014pascal}, and ADE20K \cite{zhou2017scene}. Without ImageNet \cite{ILSVRC15} pretraining, our best model significantly outperforms FRRN-B \cite{pohlen2016full} by $8.6\%$ and GridNet \cite{fourure2017residual} by $10.9\%$ on Cityscapes test set, and performs comparably with other ImageNet-pretrained state-of-the-art models \cite{wu2016wider,zhao2017pyramid,bulo2017place,deeplabv3plus2018,chen2018searching} when also exploiting the coarse annotations on Cityscapes. Notably, our best model (without pretraining) attains the same performance as DeepLabv3+ \cite{deeplabv3plus2018} (with pretraining) while being $2.23$ times faster in Multi-Adds. Additionally, our light-weight model attains the performance only $1.2\%$ lower than DeepLabv3+ \cite{deeplabv3plus2018}, while requiring $76.7\%$ fewer parameters and being $4.65$ times faster in Multi-Adds. Finally, on PASCAL VOC 2012 and ADE20K, our best model outperforms several state-of-the-art models \cite{zhou2017scene,lin2016refinenet,wu2016wider,zhao2017pyramid,xiao2018unified} while using strictly less data for pretraining.

To summarize, the contribution of our paper is four-fold:
\begin{itemize}
    \item Ours is one of the first attempts to extend NAS beyond image classification to dense image prediction.
    \item We propose a network level architecture search space that augments and complements the much-studied cell level one, and consider the more challenging joint search of network level and cell level architectures.
    \item We develop a differentiable, continuous formulation that conducts the two-level hierarchical architecture search efficiently in $3$ GPU days.
    \item Without ImageNet pretraining, our model significantly outperforms FRRN-B and GridNet, and attains comparable performance with other ImageNet-pretrained state-of-the-art models on Cityscapes. On PASCAL VOC 2012 and ADE20K, our best model also outperforms several state-of-the-art models.
\end{itemize}

\section{Related Work}
\label{sec:related_work}

\paragraph{Semantic Image Segmentation}

Convolutional neural networks \cite{lecun1989backpropagation} deployed in a fully convolutional manner (FCNs \cite{sermanet2013overfeat,long2014fully}) have achieved remarkable performance on several semantic segmentation benchmarks. Within the state-of-the-art systems, there are two essential components: multi-scale context module and neural network design. It has been known that context information is crucial for pixel labeling tasks \cite{he2004multiscale,shotton2009textonboost,kohli2009robust,ladicky2009associative,farabet2013learning,mostajabi2014feedforward,dai2015convolutional,chen2015attention}. Therefore, PSPNet \cite{zhao2017pyramid} performs spatial pyramid pooling \cite{grauman2005pyramid,lazebnik2006beyond,he2014spatial} at several grid scales (including image-level pooling \cite{liu2015parsenet}), while DeepLab \cite{chen2017deeplab,chen2017rethinking} applies several parallel atrous convolution \cite{holschneider1989real,giusti2013fast,sermanet2013overfeat,papandreou2014untangling,chen2014semantic} with different rates. On the other hand, the improvement of neural network design has significantly driven the performance from AlexNet \cite{krizhevsky2012imagenet}, VGG \cite{simonyan2014very}, Inception \cite{ioffe2015batch,szegedy2016rethinking,szegedy2017inception}, ResNet \cite{he2016deep} to more recent architectures, such as Wide ResNet \cite{Zagoruyko2016WRN}, ResNeXt \cite{xie2017aggreated}, DenseNet \cite{huang2017densely} and Xception \cite{chollet2016xception, dai2017coco}. In addition to adopting those networks as backbones for semantic segmentation, one could employ the encoder-decoder structures \cite{ronneberger2015u,badrinarayanan2015segnet,newell2016stacked,lin2016refinenet,pohlen2016full,peng2017large,islamgated,wojna2017devil,fu2017stacked,deeplabv3plus2018,zhang2018exfuse,xiao2018unified} which efficiently captures the long-range context information while keeping the detailed object boundaries. Nevertheless, most of the models require initialization from the ImageNet \cite{ILSVRC15} pretrained checkpoints except FRRN \cite{pohlen2016full} and GridNet \cite{fourure2017residual} for the task of semantic segmentation. Specifically, FRRN \cite{pohlen2016full} employs a two-stream system, where full-resolution information is carried in one stream and context information in the other pooling stream. GridNet, building on top of a  similar idea, contains multiple streams with different resolutions. In this work, we apply neural architecture search for network backbones specific for semantic segmentation. We further show state-of-the-art performance without ImageNet pretraining, and significantly outperforms FRRN \cite{pohlen2016full} and GridNet \cite{fourure2017residual} on Cityscapes \cite{Cordts2016Cityscapes}.

\vspace{-0.5\baselineskip}
\paragraph{Neural Architecture Search Method}

Neural Architecture Search aims at automatically designing neural network architectures, hence minimizing human hours and efforts.
While some works \cite{greff2015lstm, jozefowicz2015empirical, zoph2017neural, liu2018darts} search RNN cells for language tasks, more works search good CNN architectures for image classification.

Several papers used reinforcement learning (either policy gradients \cite{zoph2017neural, zoph2017learning, cai2018efficient, tan2018mnasnet} or Q-learning \cite{baker2017designing, zhong2018practical}) to train a recurrent neural network that represents a policy to generate a sequence of symbols specifying the CNN architecture.
An alternative to RL is to use evolutionary algorithms (EA), that ``evolves'' architectures by mutating the best architectures found so far \cite{real2017large, xie2017genetic, miikkulainen2017evolving, liu2018hierarchical, real2018regularized}.
However, these RL and EA methods tend to require massive computation during the search, usually thousands of GPU days. 
PNAS \cite{liu2018progressive} proposed a progressive search strategy that markedly reduced the search cost while maintaining the quality of the searched architecture.
NAO \cite{luo2018neural} embedded architectures into a latent space and performed optimization before decoding.
Additionally, several works \cite{pham2018efficient, liu2018darts, ahmed2018maskconnect} utilized architectural sharing among sampled models instead of training each of them individually, thereby further reduced the search cost.
Our work follows the differentiable NAS formulation \cite{shin2018differentiable, liu2018darts} and extends it into the more general hierarchical setting.

\vspace{-0.5\baselineskip}
\paragraph{Neural Architecture Search Space}

Earlier papers, \eg, \cite{zoph2017neural, real2017large}, tried to directly construct the entire network. 
However, more recent papers \cite{zoph2017learning, liu2018progressive, real2018regularized, pham2018efficient, liu2018darts} have shifted to searching the repeatable cell structure, while keeping the outer network level structure fixed by hand.
First proposed in \cite{zoph2017learning}, this strategy is likely inspired by the two-level hierarchy commonly used in modern CNNs.

Our work still uses this cell level search space to keep consistent with previous works. 
Yet one of our contributions is to propose a new, general-purpose network level search space, since we wish to jointly search across this two-level hierarchy.
Our network level search space shares a similar outlook as \cite{saxena2016convolutional}, but the important difference is that \cite{saxena2016convolutional} kept the entire ``fabrics'' with no intention to alter the architecture, whereas we associate an explicit weight for each connection and focus on decoding a \textit{single} discrete structure.
In addition, \cite{saxena2016convolutional} was evaluated on segmenting face images into $3$ classes \cite{kae2013augmenting}, whereas our models are evaluated on large-scale segmentation datasets such as Cityscapes \cite{Cordts2016Cityscapes}, PASCAL VOC 2012 \cite{everingham2014pascal}, and ADE20K \cite{zhou2017scene}.

The most similar work to ours is \cite{chen2018searching}, which also studied NAS for semantic image segmentation.
However, \cite{chen2018searching} focused on searching the much smaller Atrous Spatial Pyramid Pooling (ASPP) module using random search, whereas we focus on searching the much more fundamental network backbone architecture using more advanced and more efficient search methods. 
\section{Architecture Search Space}
\label{sec:space}

This section describes our two-level hierarchical architecture search space.
For the inner cell level (\secref{sec:cell_space}), we reuse the one adopted in \cite{zoph2017learning, liu2018progressive, real2018regularized, liu2018darts} to keep consistent with previous works.
For the outer network level (\secref{sec:network_space}), we propose a novel search space based on observation and summarization of many popular designs.

\subsection{Cell Level Search Space}
\label{sec:cell_space}

We define a \emph{cell} to be a small fully convolutional module, typically repeated multiple times to form the entire neural network.
More specifically, a cell is a directed acyclic graph consisting of $B$ blocks. 

Each \emph{block} is a two-branch structure, mapping from $2$ input tensors to $1$ output tensor.
Block $i$ in cell $l$ may be specified using a $5$-tuple $(I_1, I_2, O_1, O_2, C)$, where $I_1, I_2 \in \mathcal{I}_i^l$ are selections of input tensors, $O_1, O_2 \in \mathcal{O}$ are selections of layer types applied to the corresponding input tensor, and $C \in \mathcal{C}$ is the method used to combine the individual outputs of the two branches to form this block's output tensor, $H_i^l$.
The cell's output tensor $H^l$ is simply the concatenation of the blocks' output tensors $H_1^l, \hdots, H_B^l$ in this order.

The set of possible input tensors, $\mathcal{I}_i^l$, consists of the output of the previous cell $H^{l-1}$, the output of the previous-previous cell $H^{l-2}$, and previous blocks' output in the current cell $\{H_1^l, \hdots, H_i^l\}$.
Therefore, as we add more blocks in the cell, the next block has more choices as potential source of input.

The set of possible layer types, $\mathcal{O}$, consists of the following $8$ operators, all prevalent in modern CNNs:

\begin{minipage}{0.55\linewidth}
\small
\medbreak
\begin{itemize}[leftmargin=*, nolistsep]
\item $3\times3$ depthwise-separable conv
\item $5\times5$ depthwise-separable conv
\item $3\times3$ atrous conv with rate $2$
\item $5\times5$ atrous conv with rate $2$
\end{itemize}
\medbreak
\end{minipage}
\begin{minipage}{0.39\linewidth}
\small
\medbreak
\begin{itemize}[leftmargin=*, nolistsep]
\item $3\times3$ average pooling
\item $3\times3$ max pooling
\item skip connection
\item no connection (zero)
\end{itemize}
\medbreak
\end{minipage}

For the set of possible combination operators $\mathcal{C}$, we simply let element-wise addition to be the only choice.

\begin{figure*}[t]
    \centering
    \includegraphics[width=\linewidth]{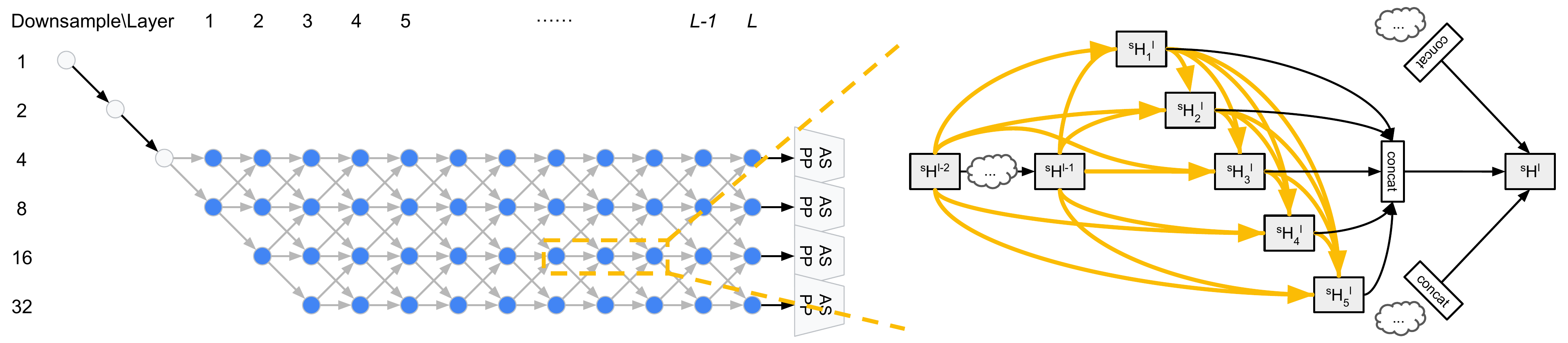}
    \caption{
    \emph{Left:} Our network level search space with $L = 12$. Gray nodes represent the fixed ``stem'' layers, and a path along the blue nodes represents a candidate network level architecture.
    \emph{Right:} During the search, each cell is a densely connected structure as described in \secref{sec:relax_cell}. Every yellow arrow is associated with the set of values $\alpha_{j\rightarrow i}$. The three arrows after \texttt{concat} are associated with $\beta_{\frac{s}{2}\rightarrow s}^l, \beta_{s\rightarrow s}^l, \beta_{2s\rightarrow s}^l$ respectively, as described in \secref{sec:relax_network}.
    Best viewed in color.}
    \label{fig:joint_space}
\end{figure*}

\begin{figure}[t]
    \centering
    \begin{subfigure}[t]{0.49\textwidth}
        \centering
        \includegraphics[width=\linewidth]{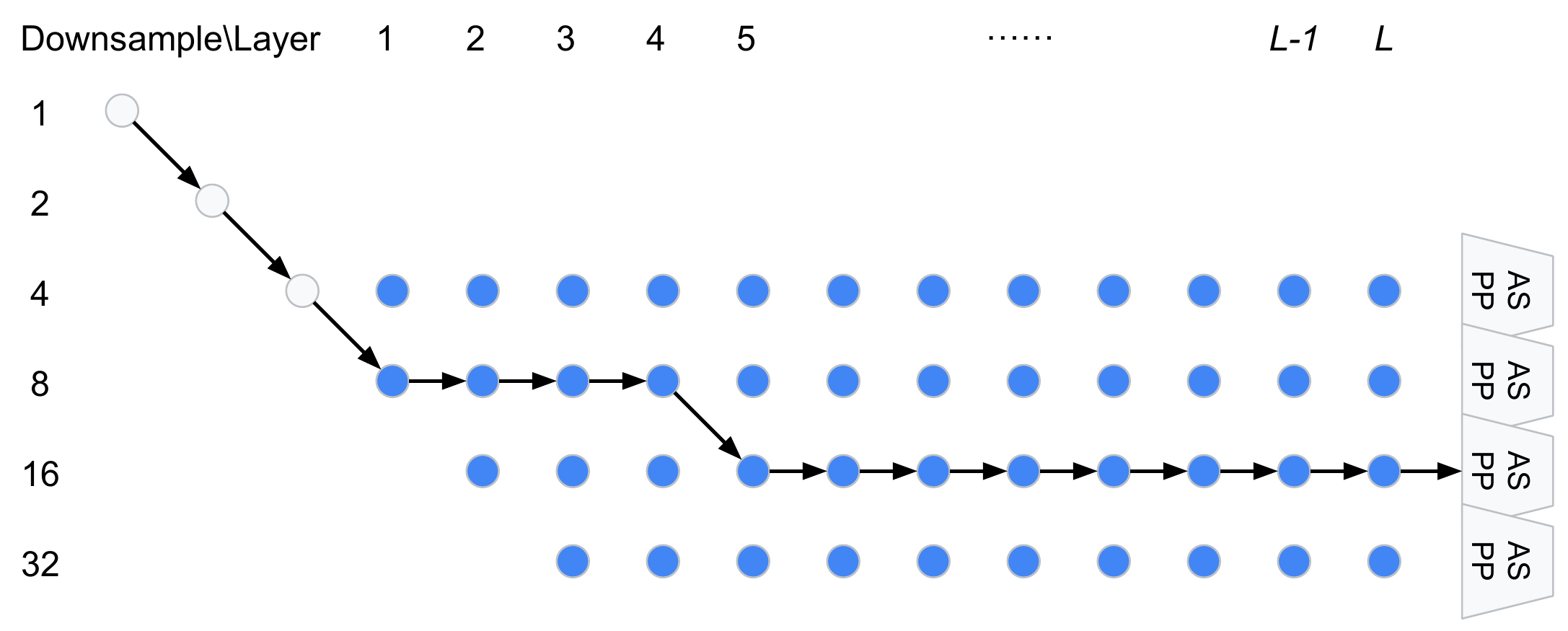}
        \caption{Network level architecture used in DeepLabv3 \cite{chen2017rethinking}.}
    \end{subfigure}
    \begin{subfigure}[t]{0.49\textwidth}
        \centering
        \includegraphics[width=\linewidth]{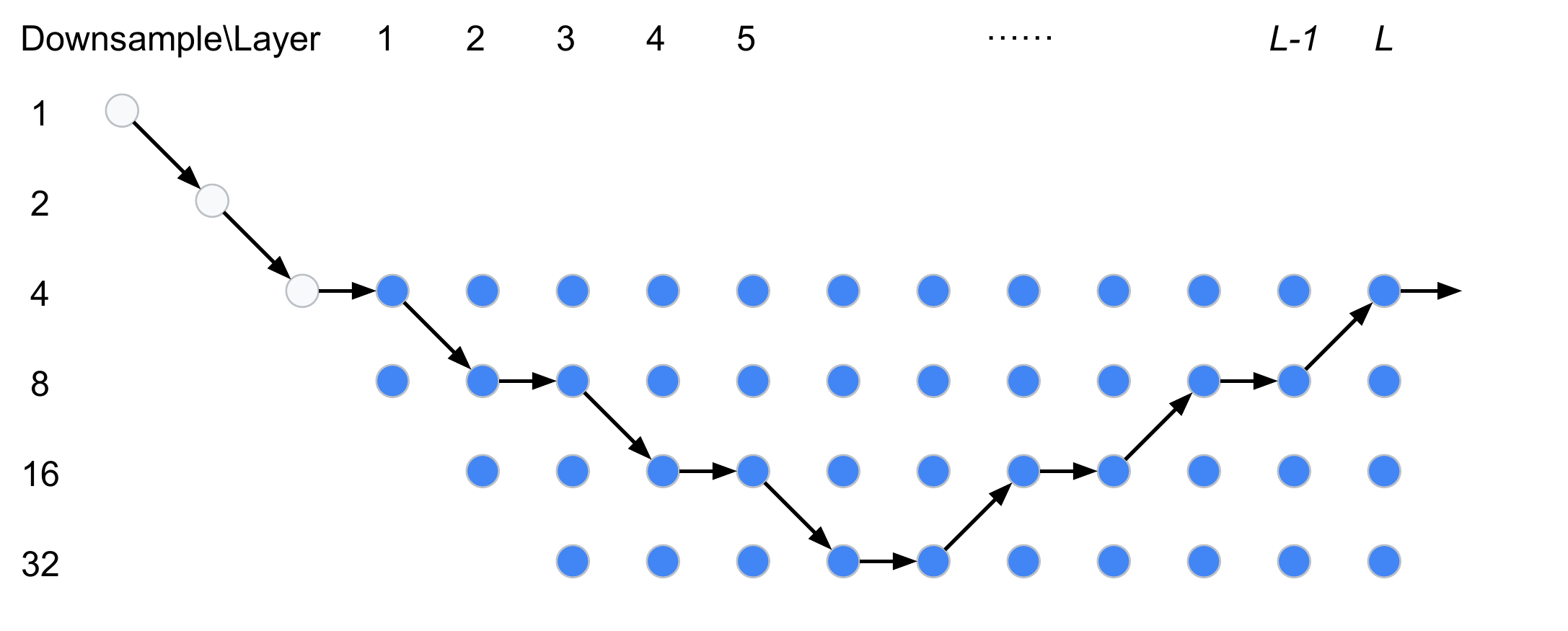}
        \caption{Network level architecture used in Conv-Deconv \cite{noh2015learning}.}
    \end{subfigure}
    \begin{subfigure}[t]{0.49\textwidth}
        \centering
        \includegraphics[width=\linewidth]{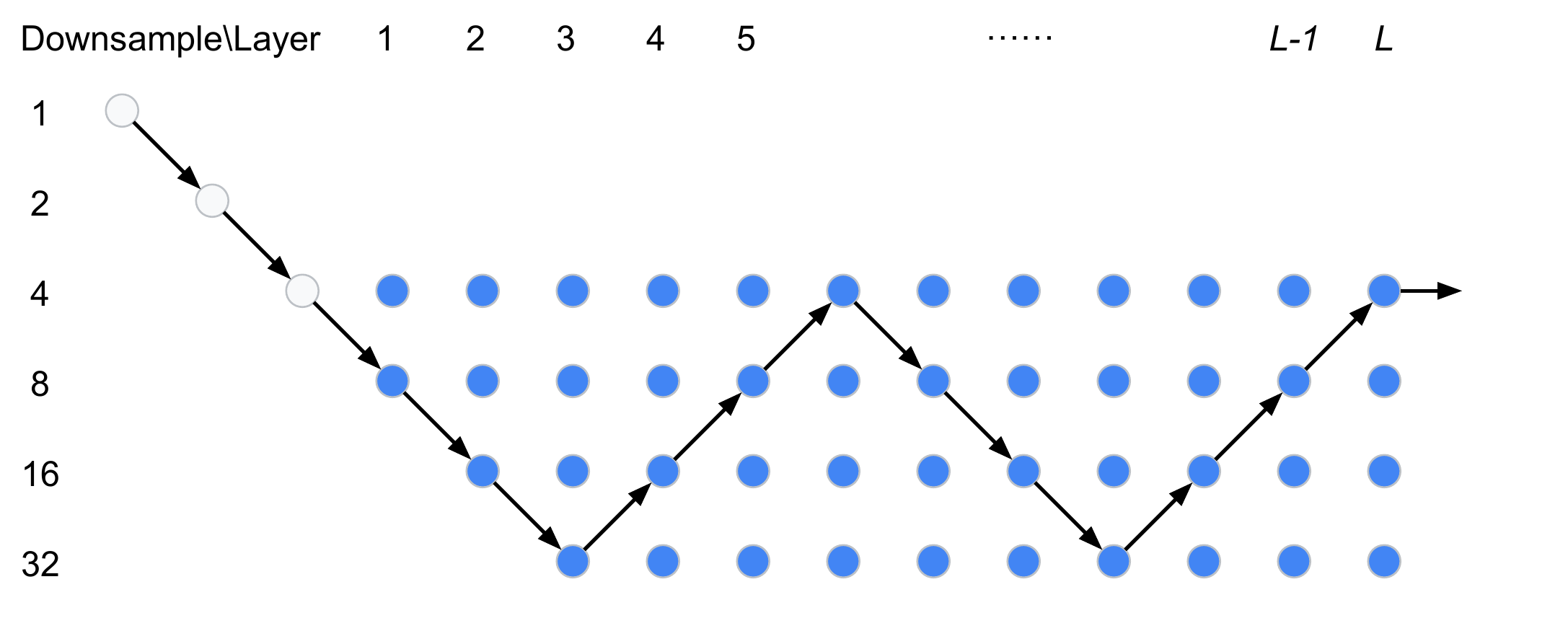}
        \caption{Network level architecture used in Stacked Hourglass \cite{newell2016stacked}.}
    \end{subfigure}
    \caption{Our network level search space is general and includes various existing designs.}
    \label{fig:popular_designs}
\end{figure}

\subsection{Network Level Search Space}
\label{sec:network_space}

In the image classification NAS framework pioneered by \cite{zoph2017learning}, once a cell structure is found, the entire network is constructed using a pre-defined pattern.
Therefore the network level was not part of the architecture search, hence its search space has never been proposed nor designed.

This pre-defined pattern is simple and straightforward: a number of ``normal cells'' (cells that keep the spatial resolution of the feature tensor) are separated equally by inserting ``reduction cells'' (cells that divide the spatial resolution by $2$ and multiply the number of filters by $2$).
This keep-downsampling strategy is reasonable in the image classification case, 
but in dense image prediction it is also important to keep high spatial resolution, and as a result there are more network level variations \cite{chen2017rethinking, noh2015learning, newell2016stacked}.

Among the various network architectures for dense image prediction, we notice two principles that are consistent:
\begin{itemize}
    \item The spatial resolution of the next layer is either twice as large, or twice as small, or remains the same.
    \item The smallest spatial resolution is downsampled by $32$.
\end{itemize}
Following these common practices, we propose the following network level search space.
The beginning of the network is a two-layer ``stem'' structure that each reduces the spatial resolution by a factor of $2$.
After that, there are a total of $L$ layers with unknown spatial resolutions, with the maximum being downsampled by $4$ and the minimum being downsampled by $32$.
Since each layer may differ in spatial resolution by at most $2$, the first layer after the stem could only be either downsampled by $4$ or $8$.
We illustrate our network level search space in \figref{fig:joint_space}.
Our goal is then to find a good path in this $L$-layer trellis.

In \figref{fig:popular_designs} we show that our search space is general enough to cover many popular designs.
In the future, we have plans to relax this search space even further to include U-net architectures \cite{ronneberger2015u, lin2016feature, shrivastava2016beyond}, where layer $l$ may receive input from one more layer preceding $l$ in addition to $l - 1$.

We reiterate that our work searches the network level architecture \emph{in addition to} the cell level architecture.
Therefore our search space is strictly more challenging and general-purpose than previous works.

\section{Methods}
\label{sec:methods}

We begin by introducing a continuous relaxation of the (exponentially many) discrete architectures that exactly matches the hierarchical architecture search described above.
We then discuss how to perform architecture search via optimization, and how to decode back a discrete architecture after the search terminates.

\subsection{Continuous Relaxation of Architectures}
\label{sec:relax}

\subsubsection{Cell Architecture}
\label{sec:relax_cell}

We reuse the continuous relaxation described in \cite{liu2018darts}.
Every block's output tensor $H_i^l$ is connected to all hidden states in $\mathcal{I}_i^l$:
\begin{equation}
    H_i^l = \sum_{H_j^l \in \mathcal{I}_i^l} O_{j\rightarrow i}(H_j^l)
    \label{eqn:cell_sum_states}
\end{equation}
In addition, we approximate each $O_{j\rightarrow i}$ with its continuous relaxation $\bar{O}_{j\rightarrow i}$, defined as:
\begin{equation}
    \bar{O}_{j\rightarrow i}(H_j^l) = \sum_{O^k \in \mathcal{O}} \alpha_{j\rightarrow i}^k O^k(H_j^l)
    \label{eqn:cell_sum_ops}
\end{equation}
where
\begin{align}
    \sum_{k=1}^{|\mathcal{O}|} &\alpha_{j\rightarrow i}^k = 1 && \forall i, j \\
    &\alpha_{j \rightarrow i}^k \geq 0 && \forall i, j, k
\end{align}
In other words, $\alpha_{j\rightarrow i}^k$ are normalized scalars associated with each operator $O^k \in \mathcal{O}$, easily implemented as softmax.

Recall from \secref{sec:cell_space} that $H^{l-1}$ and $H^{l-2}$ are always included in $\mathcal{I}_i^l$, and that $H^l$ is the concatenation of $H_1^l, \hdots, H_B^l$.
Together with \equref{eqn:cell_sum_states} and \equref{eqn:cell_sum_ops}, the cell level update may be summarized as:
\begin{equation}
    H^l = \text{Cell}(H^{l-1}, H^{l-2}; \alpha)
\end{equation}

\subsubsection{Network Architecture}
\label{sec:relax_network}

Within a cell, all tensors are of the same spatial size, which enables the (weighted) sum in \equref{eqn:cell_sum_states} and \equref{eqn:cell_sum_ops}.
However, as clearly illustrated in \figref{fig:joint_space}, tensors may take different sizes in the network level. 
Therefore in order to set up the continuous relaxation, each layer $l$ will have at most $4$ hidden states $\{{}^{4}H^l, {}^{8}H^l, {}^{16}H^l, {}^{32}H^l\}$, with the upper left superscript indicating the spatial resolution.

We design the network level continuous relaxation to exactly match the search space described in \secref{sec:network_space}.
We associated a scalar with each gray arrow in \figref{fig:joint_space}, and the network level update is:
\begin{align}
    {}^{s}H^l = &\beta_{\frac{s}{2} \rightarrow s}^l \text{Cell}({}^{\frac{s}{2}}H^{l-1}, {}^{s}H^{l-2}; \alpha) \nonumber \\
    &+ \beta_{s \rightarrow s}^l \text{Cell}({}^{s}H^{l-1}, {}^{s}H^{l-2}; \alpha) \nonumber \\
    &+ \beta_{2s \rightarrow s}^l \text{Cell}({}^{2s}H^{l-1}, {}^{s}H^{l-2}; \alpha)
    \label{eqn:network_relax}
\end{align}
where $s = 4, 8, 16, 32$ and $l = 1, 2, \hdots, L$.
The scalars $\beta$ are normalized such that
\begin{align}
    &\beta_{s \rightarrow \frac{s}{2}}^l + \beta_{s \rightarrow s}^l + \beta_{s \rightarrow 2s}^l = 1 &&\forall s, l
    \label{eqn:network_normalize} \\
    &\beta_{s \rightarrow \frac{s}{2}}^l \geq 0 \quad \beta_{s \rightarrow s}^l \geq 0 \quad \beta_{s \rightarrow 2s}^l \geq 0 &&\forall s, l
\end{align}
also implemented as softmax.

\equref{eqn:network_relax} shows how the continuous relaxations of the two-level hierarchy are weaved together. In particular, 
$\beta$ controls the outer network level, hence depends on the spatial size and layer index.
Each scalar in $\beta$ governs an entire set of $\alpha$, yet $\alpha$ specifies the same architecture that depends on neither spatial size nor layer index.

As illustrated in \figref{fig:joint_space}, Atrous Spatial Pyramid Pooling (ASPP) modules are attached to each spatial resolution at the $L$-th layer (atrous rates are adjusted accordingly).
Their outputs are bilinear upsampled to the original resolution before summed to produce the prediction.

\subsection{Optimization}
\label{sec:optimization}

The advantage of introducing this continuous relaxation is that the scalars controlling the connection strength between different hidden states are now part of the differentiable computation graph.
Therefore they can be optimized efficiently using gradient descent.
We adopt the first-order approximation in \cite{liu2018darts}, and partition the training data into two disjoint sets \textit{trainA} and \textit{trainB}.
The optimization alternates between:
\begin{enumerate}
    \item Update network weights $w$ by $\nabla_w \mathcal{L}_{trainA}(w, \alpha, \beta)$
    \item Update architecture $\alpha, \beta$ by $\nabla_{\alpha, \beta} \mathcal{L}_{trainB}(w, \alpha, \beta)$
\end{enumerate}
where the loss function $\mathcal{L}$ is the cross entropy calculated on the semantic segmentation mini-batch.
The disjoint set partition is to prevent the architecture from overfitting the training data. 

\subsection{Decoding Discrete Architectures}
\label{sec:decode}

\paragraph{Cell Architecture}

Following \cite{liu2018darts}, we decode the discrete cell architecture by first retaining the 2 strongest predecessors for each block (with the strength from hidden state $j$ to hidden state $i$ being $\max_{k, O^k \neq zero} \alpha_{j\rightarrow i}^k$; recall from \secref{sec:cell_space} that ``zero'' means ``no connection''), and then choose the most likely operator by taking the argmax.

\vspace{-\baselineskip}
\paragraph{Network Architecture}

\equref{eqn:network_normalize} essentially states that the ``outgoing probability'' at each of the blue nodes in \figref{fig:joint_space} sums to $1$. 
In fact, the $\beta$ values can be interpreted as the ``transition probability'' between different ``states'' (spatial resolution) across different ``time steps'' (layer number).
Quite intuitively, our goal is to find the path with the ``maximum probability'' from start to end.
This path can be decoded efficiently using the classic Viterbi algorithm, as in our implementation.

\section{Experimental Results}
\label{sec:experiments}
Herein, we report our architecture search implementation details as well as the search results. We then report semantic segmentation results on benchmark datasets with our best found architecture.

\begin{figure*}[t]
    \centering
    \includegraphics[width=\linewidth]{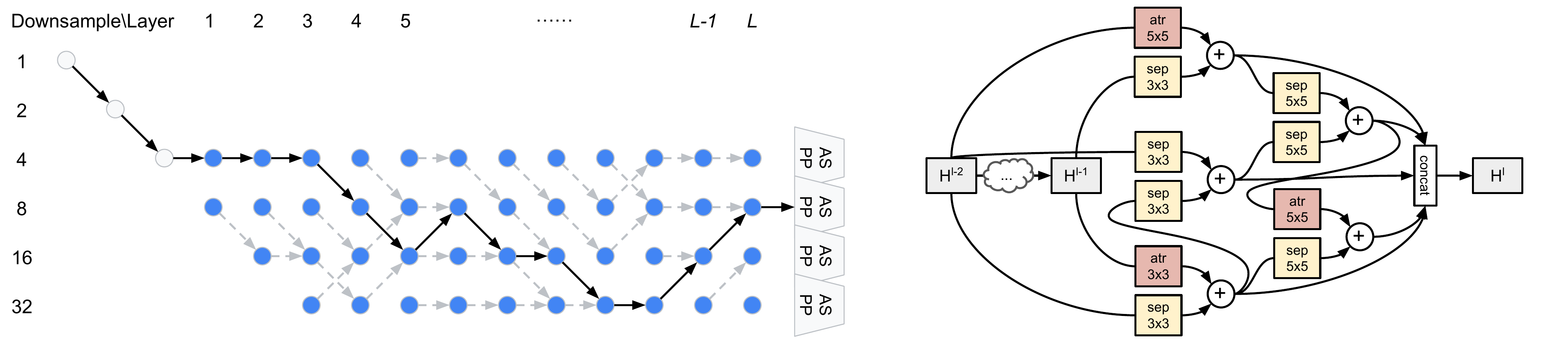}
    \caption{The Auto-DeepLab architecture found by our Hierarchical Neural Architecture Search on Cityscapes. Gray dashed arrows show the connection with maximum $\beta$ at each node. {\bf atr:} atrous convolution. {\bf sep:} depthwise-separable convolution.}
    \label{fig:hnasnet}
\end{figure*}

\subsection{Architecture Search Implementation Details}
\label{sec:search_implementation}

We consider a total of $L = 12$ layers in the network, and $B = 5$ blocks in a cell.
The network level search space has $2.9 \times 10^4$ unique paths, and the number of cell structures is $5.6 \times 10^{14}$.
So the size of the joint, hierarchical search space is in the order of $10^{19}$.

We follow the common practice of doubling the number of filters when halving the height and width of feature tensor. Every blue node in \figref{fig:joint_space} with downsample rate $s$ has $B \times F \times \frac{s}{4}$ output filters, where $F$ is the filter multiplier controlling the model capacity.
We set $F=8$ during the architecture search.
A stride $2$ convolution is used for all $\frac{s}{2}\rightarrow s$ connections, both to reduce spatial size and double the number of filters.
Bilinear upsampling followed by $1\times 1$ convolution is used for all $2s \rightarrow s$ connections, both to increase spatial size and halve the number of filters.

The Atrous Spatial Pyramid Pooling module used in \cite{chen2017rethinking} has $5$ branches: one $1\times 1$ convolution, three $3\times 3$ convolution with various atrous rates, and pooled image feature.
During the search, we simplify ASPP to have $3$ branches instead of $5$ by only using one $3 \times 3$ convolution with atrous rate $\frac{96}{s}$. 
The number of filters produced by each ASPP branch is still $B \times F \times \frac{s}{4}$.

We conduct architecture search on the Cityscapes dataset \cite{Cordts2016Cityscapes} for semantic image segmentation. 
More specifically, we use $321 \times 321$ random image crops from half-resolution ($512 \times 1024$) images in the \textit{train\_fine} set.
We randomly select half of the images in \textit{train\_fine} as \textit{trainA}, and the other half as \textit{trainB} (see \secref{sec:optimization}).

\begin{figure}[t]
    \centering
    \includegraphics[width=0.72\linewidth]{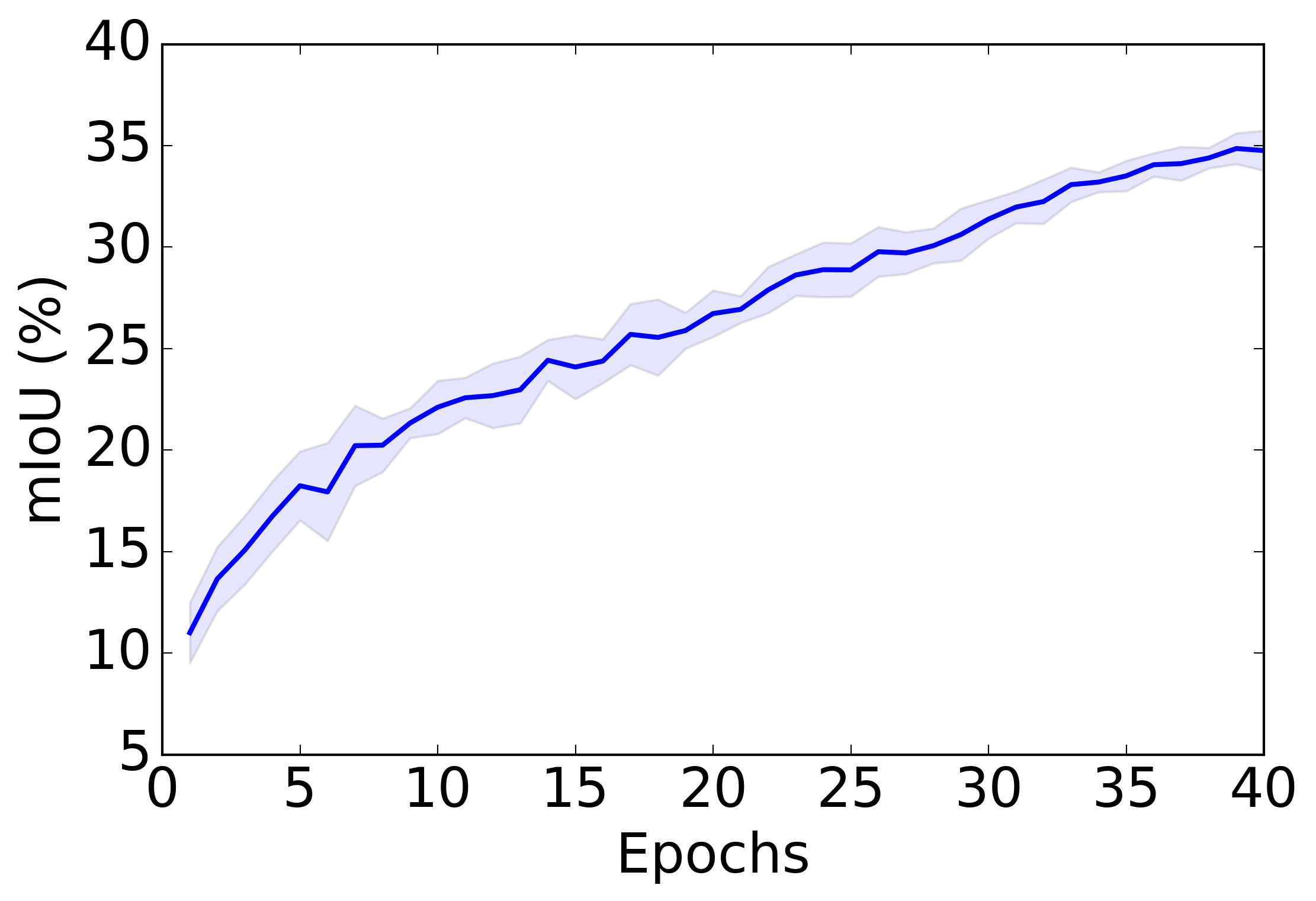}
    \vspace{-0.2cm}
    \caption{Validation accuracy during $40$ epochs of architecture search optimization across $10$ random trials. }
    \label{fig:valid_acc}
\end{figure}

The architecture search optimization is conducted for a total of $40$ epochs.
The batch size is $2$ due to GPU memory constraint.
When learning network weights $w$, we use SGD optimizer with momentum $0.9$, cosine learning rate that decays from $0.025$ to $0.001$, and weight decay $0.0003$.
The initial values of $\alpha, \beta$ before softmax are sampled from a standard Gaussian times $0.001$. 
They are optimized using Adam optimizer \cite{KingmaB14} with learning rate $0.003$ and weight decay $0.001$.
We empirically found that if $\alpha, \beta$ are optimized from the beginning when $w$ are not well trained, the architecture tends to fall into bad local optima. 
Therefore we start optimizing $\alpha, \beta$ after $20$ epochs.
The entire architecture search optimization takes about $3$ days on one P100 GPU.
\figref{fig:valid_acc} shows that the validation accuracy steadily improves throughout this process.
We also tried searching for longer epochs ($60$, $80$, $100$), but did not observe benefit.

\figref{fig:hnasnet} visualizes the best architecture found.
In terms of network level architecture, higher resolution is preferred at both beginning (stays at downsample by $4$ for longer) and end (ends at downsample by $8$).
We also show the strongest outgoing connection at each node using gray dashed arrows. 
We observe a general tendency to downsample in the first $3/4$ layers and upsample in the last $1/4$ layers.
In terms of cell level architecture, the conjunction of atrous convolution and depthwise-separable convolution is often used, suggesting that the importance of context has been learned.
Note that atrous convolution is rarely found to be useful in cells for image classification\footnote{Among NASNet-\{A, B, C\}, PNASNet-\{1, 2, 3, 4, 5\}, AmoebaNet-\{A, B, C\}, ENAS, DARTS, atrous convolution was used only once in AmoebaNet-B reduction cell.}.

\subsection{Semantic Segmentation Results}
We evaluate the performance of our found best architecture (\figref{fig:hnasnet}) on Cityscapes \cite{Cordts2016Cityscapes}, PASCAL VOC 2012 \cite{everingham2014pascal}, and ADE20K \cite{zhou2017scene} datasets.

We follow the same training protocol in \cite{chen2017rethinking, deeplabv3plus2018}. In brief, during training we adopt a polynomial learning rate schedule \cite{liu2015parsenet} with initial learning rate $0.05$, and large crop size (\eg, $769\times769$ on Cityscapes, and $513\times513$ on PASCAL VOC 2012 and resized ADE20K images). Batch normalization parameters \cite{ioffe2015batch} are fine-tuned during training. The models are trained from scratch with 1.5M iterations on Cityscapes, 1.5M iterations on PASCAL VOC 2012, and 4M iterations on ADE20K, respectively.

\begin{table}[!t]
  \centering
  \scalebox{0.72}{
  \begin{tabular}{c c | c c c | c}
    \toprule[0.2em]
    Method & ImageNet & $F$ & Multi-Adds & Params & mIOU (\%) \\
    \toprule[0.2em]
    Auto-DeepLab-S & & 20 & 333.25B & 10.15M & 79.74 \\
    Auto-DeepLab-M & & 32 & 460.93B & 21.62M & 80.04 \\
    Auto-DeepLab-L & & 48 & 695.03B & 44.42M & 80.33 \\
    \midrule
    FRRN-A \cite{pohlen2016full} & & - & - & 17.76M & 65.7 \\
    FRRN-B \cite{pohlen2016full} & & - & - & 24.78M & - \\
    DeepLabv3+ \cite{deeplabv3plus2018} & \cmark & - & 1551.05B & 43.48M & 79.55 \\
    \bottomrule[0.1em]
  \end{tabular}
  }
  \caption{Cityscapes validation set results with different Auto-DeepLab model variants. $F$: the filter multiplier controlling the model capacity. All our models are trained from \textit{scratch} and with \textit{single-scale} input during inference.}
  \label{tab:cityscapes_val}
\end{table}

\begin{table}[t!]
  \centering
  \scalebox{0.78}{
  \begin{tabular}{c | c c c c | c}
    \toprule[0.2em]
    Method & itr-500K & itr-1M & itr-1.5M & {\bf SDP} & mIOU (\%) \\
    \toprule[0.2em]
    Auto-DeepLab-S & \cmark  &        &        &        & 75.20 \\
    Auto-DeepLab-S &         & \cmark &        &        & 77.09 \\
    Auto-DeepLab-S &         &        & \cmark &        & 78.00 \\
    Auto-DeepLab-S &         &        & \cmark & \cmark & 79.74 \\
    \bottomrule[0.1em]
  \end{tabular}
  }
  \caption{Cityscapes validation set results. We experiment with the effect of adopting different training iterations (500K, 1M, and 1.5M iterations) and the Scheduled Drop Path method ({\bf SDP}). All models are trained from scratch.}
  \label{tab:cityscapes_val_2}
\end{table}

We adopt the simple encoder-decoder structure similar to DeepLabv3+ \cite{deeplabv3plus2018}. Specifically, our encoder consists of our found best network architecture augmented with the ASPP module \cite{chen2017deeplab,chen2017rethinking}, and our decoder is the same as the one in DeepLabv3+ which recovers the boundary information by exploiting the low-level features that have downsample rate $4$. Additionally, we redesign the ``stem'' structure with three $3\times3$ convolutions (with stride $2$ in the first and third convolutions). The first two convolutions have $64$ filters while the third convolution has $128$ filters. This ``stem'' has been shown to be effective for segmentation in \cite{zhao2017pyramid,wang2017understanding}.


\vspace{-0.25cm}
\subsubsection{Cityscapes}
\vspace{-0.1cm}

Cityscapes \cite{Cordts2016Cityscapes} contains high quality pixel-level annotations of $5000$ images with size $1024\times 2048$ ($2975$, $500$, and $1525$ for the training, validation, and test sets respectively) and about $20000$ coarsely annotated training images. Following the evaluation protocol \cite{Cordts2016Cityscapes}, $19$ semantic labels are used for evaluation without considering the void label.

In \tabref{tab:cityscapes_val}, we report the Cityscapes validation set results. Similar to MobileNets \cite{howard2017mobilenets,mobilenetv22018}, we adjust the model capacity by changing the filter multiplier $F$.  As shown in the table, higher model capacity leads to better performance at the cost of slower speed (indicated by larger Multi-Adds).

In \tabref{tab:cityscapes_val_2}, we show that increasing the training iterations from 500K to 1.5M iterations improves the performance by $2.8\%$, when employing our light-weight model variant, Auto-DeepLab-S. Additionally, adopting the Scheduled Drop Path \cite{larsson2017fractalnet,zoph2017learning} further improves the performance by $1.74\%$, reaching $79.74\%$ on Cityscapes validation set.

We then report the test set results in \tabref{tab:cityscapes_test}. Without any pretraining, our best model (Auto-DeepLab-L) significantly outperforms FRNN-B \cite{pohlen2016full} by $8.6\%$ and GridNet \cite{fourure2017residual} by $10.9\%$. With extra coarse annotations, our model Auto-DeepLab-L, without pretraining on ImageNet \cite{ILSVRC15}, achieves the test set performance of $82.1\%$, outperforming PSPNet \cite{zhao2017pyramid} and Mapillary \cite{bulo2017place}, and attains the same performance as DeepLabv3+ \cite{deeplabv3plus2018} while requiring $55.2\%$ fewer Mutli-Adds computations. Notably, our light-weight model variant, Auto-DeepLab-S, attains $80.9\%$ on the test set, comparable to PSPNet, while using merely 10.15M parameters and 333.25B Multi-Adds.

\begin{table}[!t]
  \centering
  \scalebox{0.9}{
  \begin{tabular}{c c c | c}
    \toprule[0.2em]
    Method & {\bf ImageNet} & {\bf Coarse} & mIOU (\%) \\
    \toprule[0.2em]
    FRRN-A \cite{pohlen2016full} &      &     & 63.0 \\
    GridNet \cite{fourure2017residual}  &     &      & 69.5 \\
    FRRN-B \cite{pohlen2016full} &      &     & 71.8 \\
    \midrule
    Auto-DeepLab-S &        &            & 79.9 \\
    Auto-DeepLab-L &        &            & 80.4 \\
    \midrule
    Auto-DeepLab-S &        & \cmark & 80.9 \\
    Auto-DeepLab-L &        & \cmark & 82.1 \\
    \midrule
    ResNet-38 \cite{wu2016wider} & \cmark & \cmark & 80.6 \\
    PSPNet \cite{zhao2017pyramid} & \cmark & \cmark & 81.2 \\
    Mapillary \cite{bulo2017place} & \cmark & \cmark & 82.0 \\
    DeepLabv3+ \cite{deeplabv3plus2018} & \cmark & \cmark & 82.1 \\
    DPC \cite{chen2018searching} & \cmark & \cmark & 82.7 \\
    DRN\_CRL\_Coarse \cite{zhuang2018dense} & \cmark & \cmark & 82.8 \\
    \bottomrule[0.1em]
  \end{tabular}
  }
  \caption{Cityscapes test set results with \textit{multi-scale} inputs during inference. {\bf ImageNet:} Models pretrained on ImageNet. {\bf Coarse:} Models exploit coarse annotations.}
  \label{tab:cityscapes_test}
\end{table}

\vspace{-0.25cm}
\subsubsection{PASCAL VOC 2012}
\vspace{-0.1cm}

PASCAL VOC 2012 \cite{everingham2014pascal} contains $20$ foreground object classes and one background class. We augment the original dataset with the extra annotations provided by \cite{hariharan2011semantic},
resulting in $10582$ (\textit{train\_aug}) training images. 

In \tabref{tab:pascal_val_2}, we report our validation set results. Our best model, Auto-DeepLab-L, with single scale inference significantly outperforms \cite{ghiasi2018dropblock} by $20.36\%$. Additionally, for all our model variants, adopting multi-scale inference improves the performance by about $1\%$. Further pretraining our models on COCO \cite{lin2014microsoft} for 4M iterations improves the performance significantly.

Finally, we report the PASCAL VOC 2012 test set result with our COCO-pretrained model variants in \tabref{tab:pascal_test}. As shown in the table, our best model attains the performance of $85.6\%$ on the test set, outperforming RefineNet \cite{lin2016refinenet} and PSPNet \cite{zhao2017pyramid}. Our model is lagged behind the top-performing DeepLabv3+ \cite{deeplabv3plus2018} with Xception-65 as network backbone by $2.2\%$. We think that PASCAL VOC 2012 dataset is too small to train models from scratch and pretraining on ImageNet is still beneficial in this case.

\begin{table}[!t]
  \centering
  \scalebox{1}{
  \begin{tabular}{c | c c | c}
    \toprule[0.2em]
    Method & {\bf MS} & {\bf COCO} & mIOU (\%) \\
    \toprule[0.2em]
    DropBlock \cite{ghiasi2018dropblock} &    &  & 53.4 \\
    \midrule \midrule
    Auto-DeepLab-S &         &        & 71.68 \\
    Auto-DeepLab-S & \cmark  &        & 72.54 \\
    \midrule
    Auto-DeepLab-M &         &        & 72.78 \\
    Auto-DeepLab-M & \cmark  &        & 73.69 \\
    \midrule
    Auto-DeepLab-L &         &        & 73.76 \\
    Auto-DeepLab-L & \cmark  &        & 75.26 \\
    \midrule \midrule
    Auto-DeepLab-S &         & \cmark & 78.31 \\
    Auto-DeepLab-S & \cmark  & \cmark & 80.27 \\
    \midrule
    Auto-DeepLab-M &         & \cmark & 79.78 \\
    Auto-DeepLab-M & \cmark  & \cmark & 80.73 \\
    \midrule
    Auto-DeepLab-L &         & \cmark & 80.75 \\
    Auto-DeepLab-L & \cmark  & \cmark & 82.04 \\
    \bottomrule[0.1em]
  \end{tabular}
  }
  \caption{PASCAL VOC 2012 validation set results. We experiment with the effect of adopting \textit{multi-scale} inference ({\bf MS}) and COCO-pretrained checkpoints ({\bf COCO}). Without any pretraining, our best model (Auto-DeepLab-L) outperforms DropBlock by $20.36\%$. All our models are not pretrained with ImageNet images.}
  \label{tab:pascal_val_2}
\end{table}

\begin{table}[!t]
  \centering
  \scalebox{0.95}{
  \begin{tabular}{c | c c | c}
    \toprule[0.2em]
    Method & {\bf ImageNet} & {\bf COCO} & mIOU (\%) \\
    \toprule[0.2em]
    \href{http://host.robots.ox.ac.uk:8080/anonymous/AGUWRT.html}{Auto-DeepLab-S} & & \cmark& 82.5 \\
    \href{http://host.robots.ox.ac.uk:8080/anonymous/8NFFR1.html}{Auto-DeepLab-M} & & \cmark & 84.1 \\
    \href{http://host.robots.ox.ac.uk:8080/anonymous/PJ8WHS.html}{Auto-DeepLab-L} & & \cmark &85.6\\
    \midrule \midrule
    RefineNet \cite{lin2016refinenet} & \cmark & \cmark & 84.2 \\
    ResNet-38 \cite{wu2016wider} & \cmark & \cmark & 84.9 \\
    PSPNet \cite{zhao2017pyramid} & \cmark & \cmark & 85.4 \\
    DeepLabv3+ \cite{deeplabv3plus2018} & \cmark & \cmark & 87.8 \\
    MSCI \cite{lin2018msci} & \cmark & \cmark & 88.0 \\
    \bottomrule[0.1em]
  \end{tabular}
  }
  \caption{PASCAL VOC 2012 test set results. Our Auto-DeepLab-L attains comparable performance with many state-of-the-art models which are pretrained on both {\bf ImageNet} and {\bf COCO} datasets. We refer readers to the official leader-board for other state-of-the-art models.}
  \label{tab:pascal_test}
\end{table}

\begin{table}[!t]
  \centering
  \scalebox{0.62}{
  \begin{tabular}{c c | c c c}
    \toprule[0.2em]
    Method & {\bf ImageNet} & mIOU (\%) & Pixel-Acc (\%) & Avg (\%)\\
    \toprule[0.2em]
    Auto-DeepLab-S &      & 40.69 & 80.60 & 60.65 \\
    Auto-DeepLab-M &      & 42.19 & 81.09 & 61.64 \\
    Auto-DeepLab-L &      & 43.98 & 81.72 & 62.85 \\
    \midrule
    CascadeNet (VGG-16) \cite{zhou2017scene} & \cmark & 34.90 & 74.52 & 54.71 \\
    RefineNet (ResNet-152) \cite{lin2016refinenet} & \cmark & 40.70 & - & - \\
    UPerNet (ResNet-101) \cite{xiao2018unified} $\dagger$ & \cmark & 42.66 & 81.01 & 61.84 \\
    PSPNet (ResNet-152) \cite{zhao2017pyramid} & \cmark & 43.51 & 81.38 & 62.45 \\
    PSPNet (ResNet-269) \cite{zhao2017pyramid} & \cmark & 44.94 & 81.69 & 63.32 \\
    DeepLabv3+ (Xception-65) \cite{deeplabv3plus2018} $\dagger$ & \cmark & 45.65 & 82.52 & 64.09 \\
    \bottomrule[0.1em]
  \end{tabular}
  }
  \caption{ADE20K validation set results. We employ \textit{multi-scale} inputs during inference. $\dagger$: Results are obtained from their up-to-date model zoo websites respectively. {\bf ImageNet:} Models pretrained on ImageNet. Avg: Average of mIOU and Pixel-Accuracy.}
  \label{tab:ade20k_val}
  \vspace{-0.25cm}
\end{table}

\subsubsection{ADE20K}
\vspace{-0.1cm}

ADE20K \cite{zhou2017scene} has $150$ semantic classes and high quality annotations of $20000$ training images and $2000$ validation images. In our experiments, the images are all resized so that the longer side is $513$ during training.

In \tabref{tab:ade20k_val}, we report our validation set results. Our models outperform some state-of-the-art models, including RefineNet \cite{lin2016refinenet}, UPerNet \cite{xiao2018unified}, and PSPNet (ResNet-152) \cite{zhao2017pyramid}; however, without any ImageNet \cite{ILSVRC15} pretraining, our performance is lagged behind the latest work of \cite{deeplabv3plus2018}.

\section{Conclusion}
\label{sec:conclusion}

In this paper, we present one of the first attempts to extend Neural Architecture Search beyond image classification to dense image prediction problems.
Instead of fixating on the cell level, we acknowledge the importance of spatial resolution changes, and embrace the architectural variations by incorporating the network level into the search space. 
We also develop a differentiable formulation that allows efficient (about $1000 \times$ faster than DPC \cite{chen2018searching}) architecture search over our two-level hierarchical search space. 
The result of the search, Auto-DeepLab, is evaluated by training on benchmark semantic segmentation datasets from scratch.
On Cityscapes, Auto-DeepLab significantly outperforms the previous state-of-the-art by $8.6\%$, and performs comparably with ImageNet-pretrained top models when exploiting the coarse annotations.
On PASCAL VOC 2012 and ADE20K, Auto-DeepLab also outperforms several ImageNet-pretrained state-of-the-art models.

For future work, within the current framework, related applications such as object detection should be plausible; we could also try untying the cell architecture $\alpha$ across different layers (\cf \cite{tan2018mnasnet}) with little computation overhead.
Beyond the current framework, a more general network level search space should be beneficial (\cf \secref{sec:network_space}).


\vspace{-0.5cm}
\ifcvprfinal
\paragraph{Acknowledgments}
We thank Sergey Ioffe for valuable feedback; Cloud AI and Mobile Vision team for support.
CL and AY acknowledge a gift from YiTu. 
\fi

{\small
\bibliographystyle{ieee}
\bibliography{egbib}
}

\end{document}